\documentclass[11pt,a4paper]{article}
\usepackage[margin=25mm]{geometry}
\usepackage[T1]{fontenc}
\usepackage{lmodern,microtype,amsmath,amssymb,booktabs,array,graphicx,caption,subcaption,float,xcolor}
\usepackage[hidelinks]{hyperref}
\usepackage[section]{placeins}
\graphicspath{{figures/publication/}}
\newcommand{\fig}[3]{\begin{figure}[htbp]\centering\includegraphics[width=.94\linewidth,height=.48\textheight,keepaspectratio]{#1}\caption{#2}\label{#3}\end{figure}}
\title{From Slices to Volumes: Comparing 2D, 2.5D, and 3D Residual Networks for Pulmonary Nodule Malignancy-Risk Classification}
\author{Bahodir Nazarov\textsuperscript{1}\thanks{Corresponding author: \href{mailto:bahodirnazarov@stu.scu.edu.cn}{bahodirnazarov@stu.scu.edu.cn}.}, Zineb Gharib\textsuperscript{1},\\ and Fatima Zazoul\textsuperscript{1}\\[5pt]\textsuperscript{1}Sichuan University, Chengdu, Sichuan, China}
\date{}
\begin{document}
\maketitle
\begin{abstract}
\textbf{Purpose:} To compare 2D, 2.5D, and 3D residual networks for pulmonary nodule malignancy-risk classification and evaluate momentum-contrast pretraining, mild augmentation, and exclusion of indeterminate ratings.
\textbf{Methods:} A retrospectively prepared LIDC-IDRI cohort contained 2,654 nodules labeled benign, indeterminate, or malignant using median radiologist ratings. Patient-disjoint splits comprised 1,876 training, 382 validation, and 396 test nodules. Representations included one axial slice, five neighboring slices, and a local CT volume. Each track evaluated a randomly initialized baseline, momentum-contrast pretraining followed by fine-tuning, and supervised training with mild augmentation. Checkpoints were selected by validation Macro-F1. The 2D experiments used three seeds; 2.5D and 3D results were single-run estimates. A separate binary experiment excluded indeterminate nodules.
\textbf{Results:} Augmented models achieved test Macro-F1 values of $0.6134\pm0.0075$ for 2D, $0.6537$ for 2.5D, and $0.5403$ for 3D; the 2D uncertainty denotes population standard deviation across seeds. Corresponding accuracies were $0.6128$, $0.6869$, and $0.5455$. Augmentation improved Macro-F1 within every track, whereas the tested pretraining recipes reduced it. Binary Macro-F1 values were $0.8227$, $0.8664$, and $0.7554$, respectively.
\textbf{Conclusion:} Augmented 2.5D achieved the highest observed performance. Differences in preprocessing, optimization, capacity, and replication limit attribution to dimensionality. Labels reflect radiologist-assessed malignancy risk rather than pathology-confirmed cancer. Code and experimental results are available at \url{https://github.com/CT-DeepLearning-Team/CT-2D-2.5D-3D-DeepLearning}.
\end{abstract}
\noindent\textbf{Keywords:} pulmonary nodules; computed tomography; deep learning; self-supervised learning.

\section{Introduction}
CT represents anatomy as an ordered stack of slices. A pulmonary nodule can therefore be represented by a single axial image, a short stack of neighboring images, or a local three-dimensional volume. These alternatives provide different spatial information and impose different learning requirements. A single image may omit through-plane variation; a neighboring-slice representation supplies limited depth context through input channels; volumetric convolutions can learn features across all three spatial axes. Whether the richer representation yields better held-out performance remains an empirical question.

LIDC-IDRI provides thoracic CT images and multiple-reader nodule annotations, including subjective malignancy ratings \cite{armato}. These ratings support research into image-based risk characterization, but their interpretation differs from a diagnosis established by histopathology. Accordingly, this study uses the term \emph{malignancy-risk classification} throughout. Its three operational classes summarize reader assessments rather than confirmed disease status.

Residual networks provide a practical common architectural family for the comparison \cite{resnet}. Self-supervised momentum contrast offers a way to initialize encoders using a larger pool of training-patient nodules without consuming malignancy labels \cite{moco,mocov2}. We also examine whether mild transformations during supervised training reduce the mismatch between fitting the available training samples and generalizing to unseen patients. Our contributions are: (i) a documented comparison of three spatial representations on one prepared cohort and fixed patient assignments; (ii) within-track comparisons of direct supervision, contrastive initialization, and mild augmentation; and (iii) an error analysis and supplementary binary experiment that retain the distinction between uncertain ratings and the easier two-class endpoint.

The study is a comparison of implemented model tracks, rather than a strict single-variable ablation. We report their differences explicitly. This qualification matters because training settings, additional intensity standardization, slice centering, parameter counts, and replication are not identical across representations.

\section{Related Work}
Prior pulmonary nodule studies have investigated different ways of incorporating local morphology and context. Shen et al. proposed a multi-crop CNN for malignancy suspiciousness classification using multi-scale nodule features \cite{shen}. Their use of multiple crops addresses scale and spatial coverage; our neighboring-slice representation instead supplies a fixed short axial stack as channels. These are related approaches to contextual information but are not interchangeable representations.

Setio et al. used multi-view convolutional networks for false-positive reduction in a pulmonary nodule detection system \cite{setio}. That work illustrates how complementary views can support nodule analysis, but its detection endpoint differs from assigning risk categories to an already identified nodule. It therefore cannot provide a directly comparable baseline for the present Macro-F1 values. Published scores obtained under different labels, cohort selection, and evaluation partitions do not provide directly comparable baselines for this study.

Recent work further combines spatial context with attention and auxiliary supervision. Zhu and Fei's Cross-ViT combines convolutional and transformer features using cross-attention, with three neighboring slices as input \cite{crossvit}. Their binary experiment uses a screened LUNA16 cohort and removes indeterminate nodules; its reader-based labeling policy differs from our median-rating classes. Thus, a short slice stack is also compatible with more complex architectures, but published binary performance does not establish performance on our three-class endpoint.

Jin et al.'s multitask Swin Transformer jointly predicts malignancy categories and nodule characteristics, and uses U-Net GAN augmentation \cite{mtst}. Its binary and multilevel experiments exclude rating 3, whereas indeterminate risk remains the largest class in our primary cohort. Liu et al.'s LNMSNet uses a 3D ResNet-18 backbone with multiscale attention and joint segmentation--classification training \cite{lnmsnet}. Its LIDC-IDRI experiment also excludes rating 3; independent hospital cohorts provide an additional evaluation setting. These studies motivate attention, auxiliary tasks, and external evaluation as relevant follow-up directions, without resolving which representation performs best under our current label and split definitions.

\begin{table}[htbp]\centering\small
\caption{Qualitative context from recent pulmonary nodule studies. These methods were not reimplemented or evaluated on our test cohort; this table is a literature comparison, not an experimental leaderboard.}\label{tab:recent}
\begin{tabular}{>{\raggedright\arraybackslash}p{.21\linewidth}>{\raggedright\arraybackslash}p{.32\linewidth}>{\raggedright\arraybackslash}p{.38\linewidth}}\toprule
Study & Representation and learning approach & Main difference from our primary endpoint\\\midrule
Cross-ViT (2025) \cite{crossvit} & Three-slice CNN--transformer feature fusion & Screened LUNA16 binary cohort; different reader aggregation; indeterminate excluded\\[3pt]
MTST (2025) \cite{mtst} & Multitask Swin Transformer; nodule-characteristic targets and GAN augmentation & Binary/multilevel targets exclude rating 3; additional supervision and generated images\\[3pt]
LNMSNet (2026) \cite{lnmsnet} & Volumetric ResNet-18 with multiscale attention; segmentation and classification & Binary LIDC-IDRI target; voxel-mask supervision; separate external hospital evaluation\\[3pt]
Present study & Residual networks with one slice, five slices, or a volume; supervised and MoCo recipes & Three median-rating groups; fixed patient-disjoint partition; unequal replication across tracks\\\bottomrule
\end{tabular}
\end{table}

This study contributes a comparison of three residual-network representations with explicit attention to uncertain ratings, pretraining transfer, and supervised regularization on a common cohort. The central question is whether additional spatial information is beneficial under the evaluated data and training constraints. Its contribution is comparative evidence with documented implementation differences rather than a new network architecture.

\section{Materials and Methods}
\subsection{Study design and source provenance}
This retrospective computational study compares three implemented model tracks on a fixed LIDC-IDRI-derived cohort. Each track evaluates direct supervised learning, contrastive initialization, mild supervised augmentation, and a separate binary sensitivity analysis. Methods and results are versioned in the accompanying repository at commit \texttt{d47e716} \cite{repository}. Evaluation is internal and patient-disjoint; no external validation cohort is included. The design emphasizes comparisons within each representation while documenting the differences that limit cross-representation causal inference. Reporting distinguishes reference-standard definition, checkpoint selection, class-specific errors, and replication, consistent with the transparency principles emphasized by CLAIM \cite{claim}.

\fig{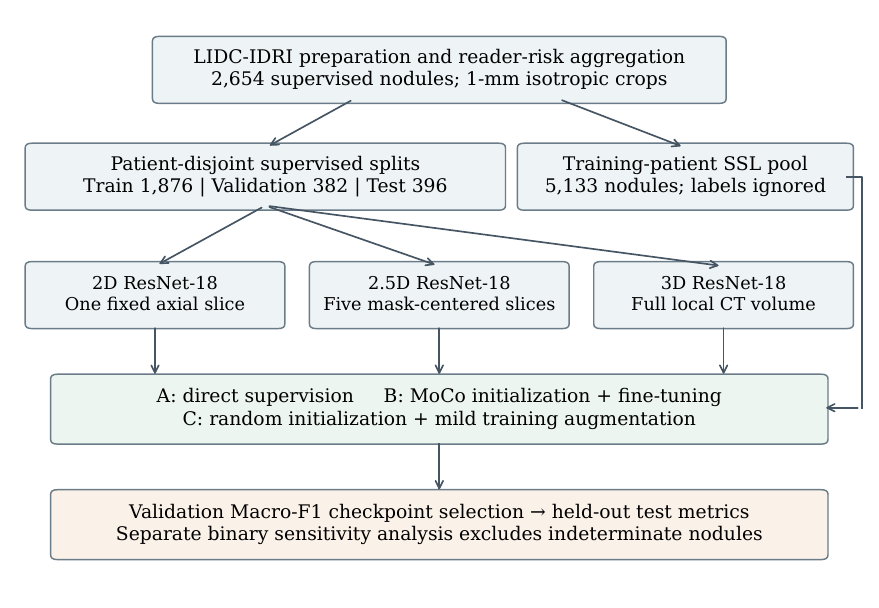}{Study design. A common prepared cohort and patient assignments feed three representation tracks. Supervised labels are benign, indeterminate, and malignant. The SSL pool includes training-assigned patients only; the binary endpoint preserves patient assignments while removing indeterminate nodules.}{fig:design}

\subsection{Dataset preparation and cohort}
The source-only preparation pipeline inventories DICOM records, selects CT series, parses XML annotations, matches image identifiers, reconstructs canonical volumes, creates reader masks and consensus nodules, resamples to isotropic spacing, extracts crops, applies technical quality control, aggregates risk ratings, and assigns patients to fixed splits. The preparation pipeline processed 1,010 patients, inventorying 244,527 DICOM records, parsing 1,318 XML files, and identifying 1,018 valid CT series. There were 7,385 consensus nodules, of which 2,672 were rated; the final supervised cohort contained 2,654 eligible nodules. These counts describe different preparation stages and should not be interpreted as independent scans or supervised examples.

The supervised cohort comprises 873 benign, 1,226 indeterminate, and 555 malignant nodules. Table~\ref{tab:cohort} gives the fixed split composition. Split separation is at the patient level: all eligible nodules from one patient receive the same assignment. The fixed split manifests record zero patient overlap among supervised splits. The SSL pool contains 5,133 nodules from 692 training-assigned patients. The SSL pool includes eligible nodules without a supervised rating and is therefore larger than the supervised training set. Validation and test patients are excluded from SSL pretraining.

\begin{table}[htbp]\centering
\caption{Supervised cohort and fixed patient-disjoint assignments. Counts are nodules except in the patient column.}\label{tab:cohort}
\begin{tabular}{lrrrrr}\toprule
Split & Patients & Benign & Indeterminate & Malignant & Total\\\midrule
Training & 602 & 622 & 876 & 378 & 1,876\\
Validation & 134 & 134 & 163 & 85 & 382\\
Test & 134 & 117 & 187 & 92 & 396\\\midrule
Total & 870 & 873 & 1,226 & 555 & 2,654\\\bottomrule
\end{tabular}
\end{table}

\subsection{Reference labels}
The median available reader malignancy rating defines the operational class: scores $1.0$--$2.0$ are benign, $2.5$--$3.0$ are indeterminate, and $3.5$--$5.0$ are malignant. Half-integer medians arise when multiple integer reader scores are aggregated. Experiments use the frozen \texttt{class\_index\_v2} metadata column instead of independently re-deriving labels. The indeterminate category is retained in the primary endpoint because it represents a substantial part of the prepared cohort and a meaningful area of reader uncertainty.

The supplementary binary experiment removes indeterminate nodules while preserving original patient assignments. It contains 1,000 training nodules (622 benign and 378 malignant), 219 validation nodules (134 and 85), and 209 test nodules (117 and 92). Binary labels use \texttt{binary\_class\_index}. The binary task is analyzed separately because both the target and the evaluated sample composition change.

\subsection{Image representations and actual preprocessing}
All source crops contain raw HU values and have shape $(104,72,80)$ in $(Z,Y,X)$ order at 1-mm isotropic spacing. Intensities are clipped and scaled as
\begin{equation}
 u(x)=\frac{\min(400,\max(-1000,x))+1000}{1400}.
\end{equation}
The common window maps the retained intensity range to $[0,1]$ and suppresses extreme values associated with acquisition and padding.

The 2D loader uses axial slice 52, producing a $1\times72\times80$ tensor. It subsequently standardizes using training-only statistics: mean approximately $0.365105$ and standard deviation $0.314459$. The implemented 2.5D loader determines the central slice from the median $z$ coordinate of positive mask voxels and extracts $z-2$ through $z+2$, producing $5\times72\times80$. It uses HU scaling without additional standardization. Masks are therefore used for 2.5D slice selection, although they are not input channels. The 3D loader supplies the complete scaled crop as $1\times104\times72\times80$, without the 2D standardization step. Validation and test inference use no stochastic augmentation.

The shared dataset and HU window do not imply identical preprocessing. Mask-dependent centering in 2.5D, fixed centering in 2D, and additional 2D standardization are retained as characteristics of the evaluated implementations and considered when interpreting cross-track differences.

\subsection{Network architectures}
Both planar tracks use randomly initialized torchvision ResNet-18 architectures with the first convolution adapted to one or five channels. The input convolution has 64 output channels, kernel size $7\times7$, stride 2, and padding 3. Native crop dimensions are retained. The 2D classifier uses a 512-dimensional pooled feature, dropout, and a linear head. The 2.5D classifier replaces the standard final layer with a three-output head. The three-class models have 11,171,779 and 11,184,323 parameters, respectively.

The final volumetric model is a lightweight 3D ResNet-18, with residual-block counts $[2,2,2,2]$, channel widths $16,32,64,128$, adaptive global-average pooling, dropout $0.20$, and a linear head. It has 2,073,747 three-class parameters. Although the volumetric network has fewer parameters, its larger input and spatial operations create a different computational and optimization problem. Equal architectural family does not imply equal capacity, receptive fields, or training cost.

\subsection{Experimental conditions and optimization}
Within each representation, Model A denotes random initialization followed by supervised training without augmentation. Model B initializes the encoder from momentum-contrast pretraining and then fine-tunes it using the corresponding nonaugmented supervised recipe. Model C starts randomly and introduces mild training-only augmentation. The binary experiment uses the Model C recipe with a two-output head. Final settings are summarized in Table~\ref{tab:training}. Figure~\ref{fig:design} outlines the cohort-to-evaluation design.

\begin{table}[htbp]\centering\small
\caption{Final supervised training settings. Track differences are retained rather than described as a common optimizer protocol.}\label{tab:training}
\begin{tabular}{p{.29\linewidth}rrr}\toprule
Setting & 2D & 2.5D & 3D\\\midrule
Optimizer & AdamW & AdamW & AdamW\\
Batch size & 64 & 8 & 4\\
Initial learning rate & $10^{-3}$ & $10^{-3}$ & $10^{-4}$\\
Weight decay & $0.05$ & $10^{-4}$ & $10^{-4}$\\
Maximum epochs & 80 & 20 & 50\\
Early stopping patience & 20 & 5 & 10\\
Loss class weights & Inverse frequency & None & Inverse frequency\\
Label smoothing & $0.10$ & None & None\\
Classifier dropout & $0.50$ & None & $0.20$\\
Supervised LR schedule & Cosine & Constant & Constant\\
Replication & Seeds 0,1,2 & Single run & Seed 42\\\bottomrule
\end{tabular}
\end{table}

The supervised optimizer is AdamW, which decouples weight decay from the adaptive gradient update \cite{adamw}. For the weighted tracks, class weights are derived only from training counts as $w_k=N/(Kn_k)$. With three classes the weights are approximately $[1.005,0.714,1.654]$. The 2.5D scripts use unweighted cross-entropy. Every track selects the checkpoint with maximum validation Macro-F1 and applies early stopping after the stated number of epochs without improvement. Test metrics are computed at the selected checkpoint. The repository describes validation-driven selection; it does not provide a complete chronology that independently establishes the number of test accesses during project development.

\subsection{Momentum-contrast pretraining}
MoCo learns representations by comparing two augmented views of the same nodule against a queue of representations from other samples \cite{moco}. For normalized query $q$, positive key $k^+$, queued keys $k_j$, and temperature $\tau$, the contrastive loss is
\begin{equation}
\mathcal L_{\mathrm{NCE}}=-\log\frac{\exp(q^\top k^+/\tau)}{\exp(q^\top k^+/\tau)+\sum_{j=1}^{Q}\exp(q^\top k_j/\tau)}.
\end{equation}
The key encoder follows the query encoder by an exponential moving average. A projection head is used during contrastive learning and is discarded when the encoder transfers to classification. The planar 2D implementation follows a MoCo v2-style recipe with an MLP head and stronger view augmentation \cite{mocov2}.

\begin{table}[htbp]\centering\small
\caption{Pretraining configurations recorded in the repository. All use momentum $0.999$ and projection output dimension 128.}\label{tab:ssl}
\begin{tabular}{lrrr}\toprule
Setting & 2D & 2.5D & 3D\\\midrule
Queue size & 4,096 & 512 & 4,096\\
Temperature & 0.20 & 0.07 & 0.07\\
Batch size & 128 & 8 & 4\\
Epochs & 200 & 20 & 50\\
Optimizer & SGD & AdamW & AdamW\\
Initial learning rate & 0.015 & 0.001 & 0.0003\\
Weight decay & 0.0001 & 0.0001 & 0.0001\\\bottomrule
\end{tabular}
\end{table}

The 2D run uses SGD momentum $0.9$ and cosine scheduling; 3D also uses cosine scheduling. The projection MLP is $512\to512\to128$ in 2D and $128\to128\to128$ in 3D; the 2.5D output dimension is 128. These substantially different pretraining budgets mean that the comparison tests specific track recipes rather than a uniform SSL configuration. The SSL pool is restricted to training-assigned patients, and risk labels are ignored during pretraining.

\subsection{Training augmentation}
The 2D mild recipe includes horizontal and vertical flips, affine rotation up to $15^\circ$, translation parameters up to $0.06$ in the implemented normalized grid, scale $0.9$--$1.1$, intensity scale $0.9$--$1.1$, intensity shift $\pm0.05$, and Gaussian noise with standard deviation $0.02$. Its affine transform is sampled with probability $0.8$; flips have probability $0.5$ each, intensity jitter $0.5$, and noise $0.3$. Augmentation occurs before standardization.

The 2.5D recipe applies the same flips and intensity/noise settings jointly across all five channels, preserving their order and correspondence. It does not use the 2D affine transform. The 3D recipe uses rotations up to $5^\circ$ around each axis, translations up to three voxels, intensity scale $0.95$--$1.05$, shift $\pm0.03$, and Gaussian noise with standard deviation $0.01$. Values are clipped back to $[0,1]$. These representation-specific recipes should be interpreted through within-track A--C contrasts.

\subsection{Evaluation and uncertainty}
The primary reported metric is test Macro-F1 after validation-based checkpoint selection. For $K$ classes,
\begin{equation}
\mathrm{Macro\!\text{-}\!F1}=\frac{1}{K}\sum_{k=1}^{K}\frac{2P_kR_k}{P_k+R_k},\qquad
\mathrm{Balanced\ accuracy}=\frac{1}{K}\sum_{k=1}^{K}R_k.
\end{equation}
We also report accuracy, per-class precision and recall, confusion matrices, and macro one-versus-rest ROC-AUC for three-class tasks. Binary AUC uses malignant-class probabilities. The majority-class test predictor has accuracy $187/396=0.4722$, balanced accuracy $1/3$, and Macro-F1 approximately $0.2138$; it contextualizes performance under class imbalance.

The 2D values aggregate three downstream seeds, while the other tracks lack corresponding repeated-seed estimates. The three 2D Model B fine-tuning runs share the single seed-0 pretrained encoder; they do not constitute three independent pretraining replications. The supplied 2D standard deviations use the population convention (divisor three for three seeds); they describe seed variation and are not confidence intervals on representation differences. Saved 2D runs additionally provide nodule-level percentile bootstrap intervals from 2,000 resamples. Because patients can contribute multiple nodules, those intervals do not fully represent within-patient dependence. No paired significance test, patient-cluster bootstrap, or pooled cross-track confidence interval is claimed.

\section{Results}
\subsection{Three-class performance}
Table~\ref{tab:results} reproduces the final machine-readable team comparison, rounded to four decimals. Model A gives similar Macro-F1 for 2D ($0.6059$) and 2.5D ($0.6075$), while 3D is lower ($0.4928$). Across Model C runs, 2.5D has the highest observed Macro-F1 ($0.6537$), accuracy ($0.6869$), balanced accuracy ($0.6413$), and ROC-AUC ($0.7894$). Figure~\ref{fig:comparison} presents the final three-class results alongside the separate binary endpoint.

\begin{table}[htbp]\centering\small
\caption{Primary three-class results. 2D values are three-seed summaries; supplied SDs apply to Macro-F1. 2.5D and 3D are single runs. AUC is macro one-versus-rest.}\label{tab:results}
\begin{tabular}{llrrrrr}\toprule
Input & Model & Val. F1 & Accuracy & Bal. acc. & Test Macro-F1 & AUC\\\midrule
2D & A & .5982 & .6002 & .6061 & $.6059\pm.0239$ & .7565\\
2D & B & .6135 & .5926 & .5908 & $.5957\pm.0086$ & .7579\\
2D & C & .5962 & .6128 & .6110 & $.6134\pm.0075$ & .7595\\\midrule
2.5D & A & .5883 & .6086 & .6060 & .6075 & .7583\\
2.5D & B & .5516 & .5126 & .4914 & .5125 & .6922\\
2.5D & C & .5857 & .6869 & .6413 & .6537 & .7894\\\midrule
3D & A & .5139 & .4949 & .4790 & .4928 & .6564\\
3D & B & .5127 & .4596 & .4683 & .4730 & .6692\\
3D & C & .5356 & .5455 & .5461 & .5403 & .7165\\\bottomrule
\end{tabular}
\end{table}
\fig{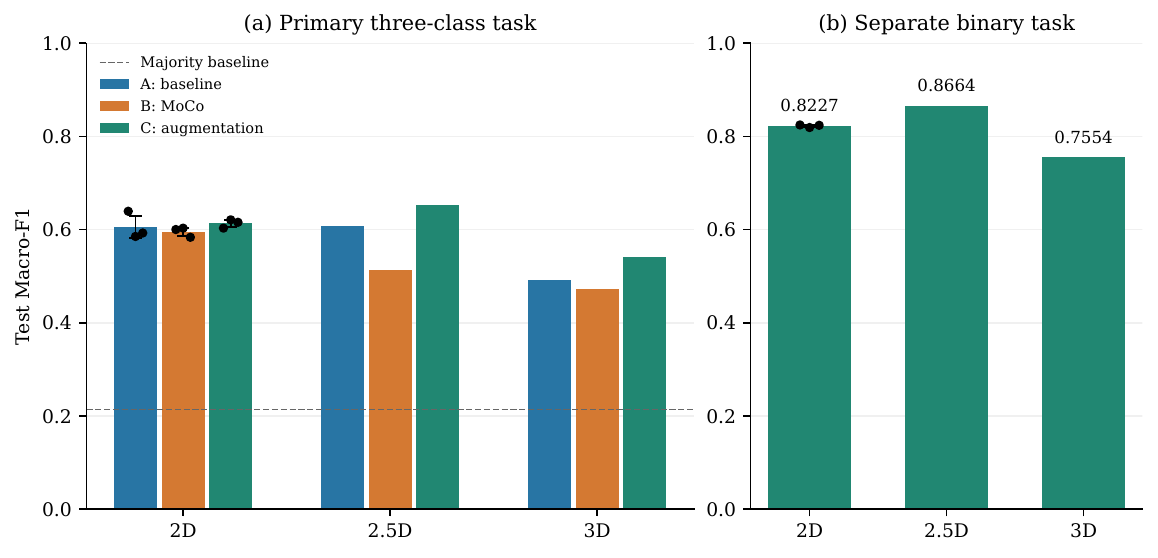}{Final test Macro-F1. (a) Three-class experiments A, B, and C; (b) separate binary experiment using the C training recipe. In 2D, bars show three-seed means, dots individual seeds, and error bars the population SD used in the stored summary. The 2.5D and 3D bars are single runs without uncertainty estimates. The dashed line in (a) is the always-indeterminate baseline. Binary and three-class scores evaluate different cohorts and targets.}{fig:comparison}

\subsection{Augmentation and contrastive initialization}
The A--C changes in test Macro-F1 are $+0.0075$ for 2D, $+0.0462$ for 2.5D, and $+0.0475$ for 3D. The positive direction is consistent across tracks; the small 2D difference should not be interpreted as established superiority given seed variation. The 2.5D report records a train--validation Macro-F1 gap of $0.2403$ for A and $0.0612$ for C at their selected checkpoints. These values support a regularization interpretation for that run, although training measurements under stochastic transformations are not identical to clean evaluation measurements.

MoCo decreases downstream Macro-F1 relative to A in all tracks: $-0.0102$, $-0.0950$, and $-0.0198$ for 2D, 2.5D, and 3D. The 2.5D pretraining loss falls from approximately $3.7805$ to $0.1156$ over 20 epochs, demonstrating optimization of the contrastive objective but not improved risk-class separation. The 3D AUC increases slightly from $0.6564$ to $0.6692$ under B while Macro-F1 decreases, illustrating that probability ranking and discrete classification need not move together.
\fig{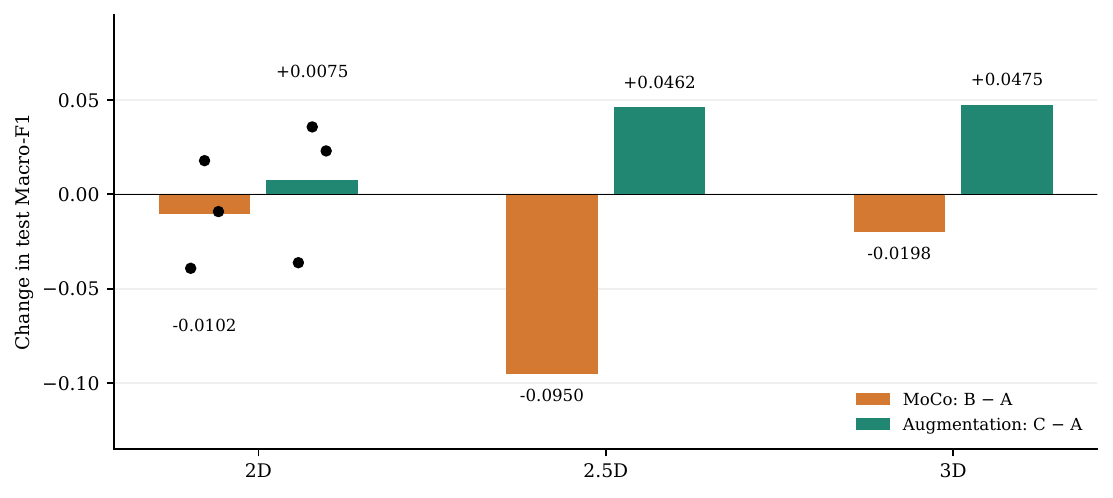}{Observed changes in test Macro-F1 compared to Model A: contrastive initialization (B--A) and mild augmentation (C--A). Bars show differences of reported three-seed means for 2D and single-run differences for 2.5D/3D; dots show same-seed downstream differences for 2D. No confidence interval or significance test for the changes is implied.}{fig:changes}

\subsection{Class-specific errors}
Table~\ref{tab:class} reports the class-specific metrics for Model C, and Figure~\ref{fig:confusion} displays the corresponding error distributions.
\begin{table}[htbp]\centering\small
\caption{Model C class-specific test performance. 2D entries average the three seed-specific metrics; 2.5D and 3D entries are single-run estimates. Class names denote radiologist risk groups.}\label{tab:class}
\begin{tabular}{llrrr}\toprule
Input & Reference class & Precision & Recall & F1\\\midrule
2D & Benign & 0.5467 & 0.5499 & 0.5421\\
2D & Indeterminate & 0.6251 & 0.6346 & 0.6274\\
2D & Malignant & 0.6950 & 0.6486 & 0.6708\\
2.5D & Benign & 0.8261 & 0.3248 & 0.4663\\
2.5D & Indeterminate & 0.6218 & 0.9144 & 0.7403\\
2.5D & Malignant & 0.8400 & 0.6848 & 0.7545\\
3D & Benign & 0.4231 & 0.3761 & 0.3982\\
3D & Indeterminate & 0.5789 & 0.5882 & 0.5836\\
3D & Malignant & 0.6078 & 0.6739 & 0.6392\\
\bottomrule\end{tabular}\end{table}

The 2.5D Model C confusion matrix contains correct counts of 38/117 benign, 171/187 indeterminate, and 63/92 malignant nodules. Thus its highest overall three-class score coexists with benign recall of only $0.3248$. Seventy-five benign nodules are predicted indeterminate, and 29 malignant nodules are also assigned to that category. Precision is $0.8261$, $0.6218$, and $0.8400$ for benign, indeterminate, and malignant classes. The model predicts indeterminate for 275/396 test nodules, substantially more than the 187 true indeterminate labels.

The 3D Model C correct counts are 44/117, 110/187, and 62/92, corresponding to recalls $0.3761$, $0.5882$, and $0.6739$. Its lower overall Macro-F1 does not mean that every class is worse: benign recall is higher than in the 2.5D run, whereas indeterminate recall is lower. Such tradeoffs are obscured by accuracy alone. The pooled 2D confusion matrix combines predictions from three evaluations of the same test cohort and must not be interpreted as 1,188 independent test nodules. Relative to the 2.5D C run, the 2D seed-averaged recall profile assigns less of its performance to the indeterminate class, reinforcing that the aggregate ranking does not determine the preferred class-specific tradeoff.
\fig{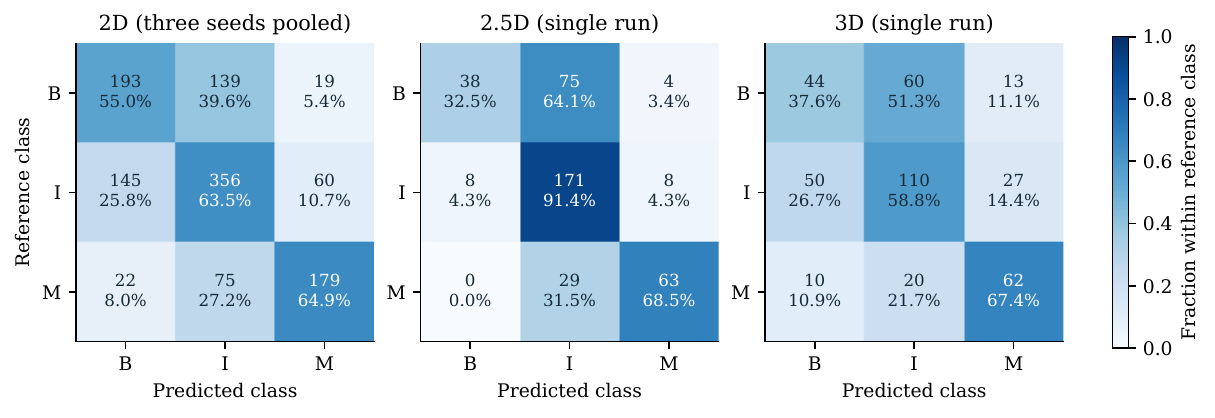}{Row-normalized Model C confusion matrices. B, I, and M denote benign, indeterminate, and malignant risk groups. Labels within cells are counts and row percentages. The 2D counts pool three evaluations of the same 396 nodules (1,188 predictions), whereas the other matrices each contain one evaluation of 396 nodules. Color encodes the fraction within a reference class, enabling comparison despite unequal prediction counts.}{fig:confusion}

\subsection{Binary sensitivity analysis}
Table~\ref{tab:binary} summarizes the separate binary endpoint. The highest observed binary Macro-F1 is $0.8664$ for 2.5D, with accuracy $0.8708$ and AUC $0.9215$. Relative to three-class Model C, the numerical Macro-F1 increases are $0.2093$, $0.2127$, and approximately $0.2151$. These are differences between tasks, not performance gains on an unchanged population. Excluding the indeterminate category removes 187 of the 396 test nodules and changes decision boundaries and chance performance.
\begin{table}[htbp]\centering\small
\caption{Supplementary benign-versus-malignant results on 209 test nodules. Original patient assignments are retained.}\label{tab:binary}
\begin{tabular}{lrrrrr}\toprule
Input & Val. F1 & Accuracy & Bal. acc. & Test Macro-F1 & AUC\\\midrule
2D & .8328 & .8293 & .8182 & $.8227\pm.0024$ & .8644\\
2.5D & .8583 & .8708 & .8614 & .8664 & .9215\\
3D & .7954 & .7608 & .7538 & .7554 & .8168\\\bottomrule
\end{tabular}
\end{table}

\subsection{Replication and learning behavior}
The final 2D C test Macro-F1 values for seeds 0, 1, and 2 are $0.6033$, $0.6211$, and $0.6158$. Their saved nodule-bootstrap 95\% intervals are $[0.5551,0.6486]$, $[0.5736,0.6679]$, and $[0.5661,0.6640]$. These intervals describe individual seed predictions and are not an interval around the three-seed mean. Corresponding baseline seed values range from $0.5854$ to $0.6395$, so the apparent A--C average improvement does not occur for every seed. Figure~\ref{fig:learning} displays the saved 2D and 3D A/C histories to distinguish training fit from validation behavior. The final 3D C checkpoint is at epoch 16; the 2.5D C checkpoint is at epoch 12.

\fig{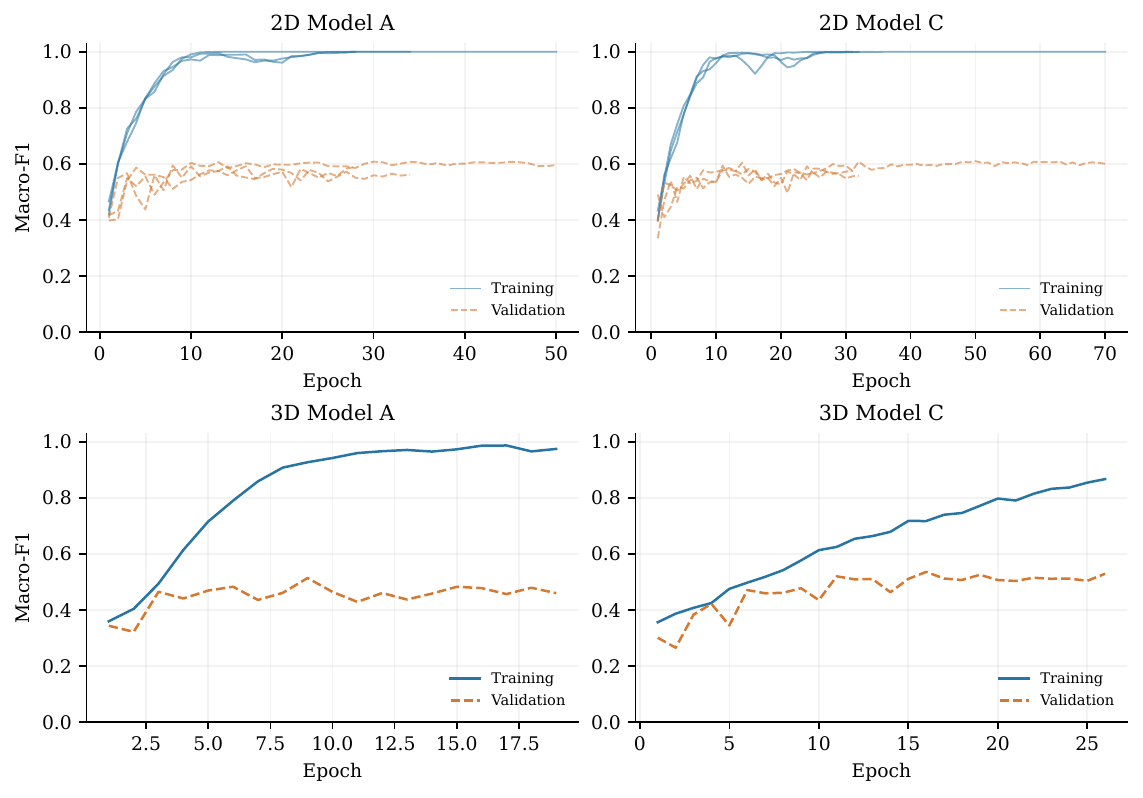}{Training and validation Macro-F1 for Models A and C. Planar 2D panels show all three seeds as separate curves; volumetric 3D panels show the single saved run. Solid curves denote training and dashed curves validation. Training metrics for C are measured with stochastic augmentation and therefore do not define a perfectly matched clean train--validation comparison. The 2.5D gap is described from its saved report rather than reconstructed as a numerical time series.}{fig:learning}

\section{Discussion}
\subsection{Spatial context and generalization}
The strongest observed three-class result is achieved by augmented 2.5D, not by the full-volume network. A short stack may provide useful local depth context while retaining planar feature extraction. However, this explanation is a hypothesis about the present implemented tracks. The baseline 2D and 2.5D scores are almost identical, and several aspects besides spatial context differ. The comparison does not establish that five slices are intrinsically optimal or that 3D information is harmful.

The 3D model sees far more input voxels while using fewer parameters and smaller batches. Its learning dynamics are affected by these design choices, normalization, and available labeled examples. More context can introduce additional variation that a limited cohort does not support learning robustly. Conversely, another 3D architecture, training budget, crop extent, or regularization recipe might produce a different ranking. The results support selecting representations empirically under the intended data budget.

\subsection{Regularization and self-supervised transfer}
The consistent positive A--C direction suggests that mild supervised augmentation is useful in these experiments. The larger observed changes in 2.5D and 3D, and the reduced reported 2.5D train--validation gap, support further study of regularization. Because transformations differ by representation and only 2D has repeated-seed results, the data do not quantify a universal augmentation effect.

The negative A--B changes are useful evidence about the specific pretraining recipes. Optimizing instance discrimination does not guarantee that features support the downstream class definitions. View strength, queue composition, optimization budget, temperature, and fine-tuning can influence transfer. Multiple nodules from one patient may also enter a queue without being recognized as related examples. This is a plausible issue for future experimentation, not a demonstrated cause of the present results. The study cannot conclude that self-supervised learning in general is ineffective.

\subsection{Uncertainty and interpretation of the endpoint}
The binary analysis indicates that the intermediate-risk category contributes to a difficult three-class decision problem. It does not isolate label ambiguity as the only explanation: the binary cohort is smaller, has a different class balance, and uses two output classes. The 2.5D C error profile further shows that high indeterminate recall can accompany low benign recall. A strong aggregate score must therefore be read together with class-specific errors.

The endpoint is agreement with an aggregated reader rating. It is not a measurement of cancer sensitivity or specificity, prospective benefit, or clinical utility. The three-class labels also have an ordering that the nominal softmax classifiers do not explicitly encode. Future work could compare ordinal objectives, retain distributions of reader ratings, or model disagreement directly. These alternatives would require new experiments rather than reinterpretation of the current saved scores.

\subsection{Limitations and next experiments}
First, this study uses one fixed patient partition and no external dataset. Acquisition variation and domain shift remain untested. Second, repeated-seed reporting is unequal, and nodules within patients are correlated. Patient-cluster uncertainty estimation and matched repeated runs are needed before claiming statistical superiority. Third, mask-based centering in 2.5D, additional 2D standardization, unequal loss weighting, and distinct optimization and augmentation recipes confound a dimensionality-only interpretation.

Fourth, early exploratory runs and parameter sweeps are retained in the repository. Although checkpoint selection uses validation Macro-F1, the committed artifacts alone cannot establish complete independence of final design decisions from prior test observations. Any future confirmatory comparison should preregister its protocol and reserve a new evaluation cohort. Fifth, the repository lacks redistributed checkpoints and imaging arrays, and some scripts contain machine-specific paths. Reproduction requires separate dataset access and path configuration. Finally, radiologist-derived labels are subjective and lack universal pathology confirmation.

A stronger follow-up would harmonize centering and intensity processing, standardize the downstream loss and selection policy, compare computational budgets, repeat every track across matched seeds, and report patient-level paired differences on an untouched evaluation set. These steps would distinguish the effect of representation from the behavior of the current three implementations.

\subsection{Scope of comparison with recent methods}
The current experiments do not include direct matched evaluations of Cross-ViT, MTST, or LNMSNet \cite{crossvit,mtst,lnmsnet}. Our three representation tracks are residual-network implementations, not three distinct state-of-the-art algorithms. Published numbers from the recent studies are not used as experimental baselines because their cohort selection, risk aggregation, excluded ratings, supervision, and test partitions differ. This is a methodological reason to avoid a numerical cross-study ranking; it does not replace evaluation of those methods on the same test patients.

A matched benchmark would require adapting their classification heads to the three median-rating classes, preserving the frozen patient manifests, defining which auxiliary annotations are available during training, and retraining under a documented selection and replication policy. The existing saved experiments do not provide those comparisons. Accordingly, our conclusions concern the reported residual-network tracks and training recipes. They neither establish competitive superiority over recent methods nor attribute the observed ranking solely to input dimensionality.

\section{Conclusion}
On the fixed LIDC-IDRI-derived patient-disjoint cohort, augmented 2.5D achieved the highest observed three-class test Macro-F1 ($0.6537$). Mild augmentation improved the reported mean or single-run Macro-F1 across all tracks, while the tested MoCo recipes did not. Removing indeterminate nodules produced substantially higher scores on a separate binary endpoint. The findings show that full-volume modeling did not outperform the planar alternatives under the evaluated data and training constraints. They motivate carefully harmonized and replicated comparisons, without establishing universal superiority of 2.5D or clinical diagnostic performance.

\section*{Author contributions}
Bahodir Nazarov led the 2D ResNet-18 track; Fatima Zazoul led the five-slice 2.5D ResNet-18 track; Zineb Gharib led the local-volume 3D ResNet-18 track.
\section*{Data and code availability}
Code, experimental reports, metrics, and figures are available in the team repository \cite{repository}. LIDC-IDRI imaging data and reader annotations are distributed separately by The Cancer Imaging Archive (TCIA) \cite{lidcdata,tcia}. The repository does not redistribute medical arrays or model checkpoints. The reported experiment summaries, seed-specific results, and figure-generating code support inspection of the analyses.
\section*{Acknowledgements}
We thank Professor Haixian Zhang of Sichuan University for his supervision and guidance throughout this project. We acknowledge the National Cancer Institute and the Foundation for the National Institutes of Health for enabling the public LIDC-IDRI resource used in this study \cite{lidcdata}.
\section*{Statements and Declarations}
\subsection*{Funding}
No funding was received for this study.
\subsection*{Competing interests}
On behalf of all authors, the corresponding author states that there is no conflict of interest.
\subsection*{Ethical considerations}
This study is a secondary analysis of publicly released LIDC-IDRI CT images and reader annotations \cite{armato,lidcdata,tcia}. No participants were recruited and no additional imaging or intervention was performed for this study.
% Submission metadata still needed: authors must confirm their institution's
% determination regarding ethics approval and consent for secondary use of LIDC-IDRI.
% No ethics approval, exemption, or consent waiver is assumed here.

\end{document}